\documentclass[10pt,twocolumn,letterpaper]{article}

\usepackage{cvpr}
\usepackage{cite}
\usepackage{times}
\pdfoutput=1
\usepackage{epsfig}
\usepackage{graphicx}
\usepackage{amsmath}
\usepackage{amssymb}
\usepackage{booktabs}
\usepackage{pdfpages}

\newcommand{\vct}[1]{\boldsymbol{#1}}

\usepackage[pagebackref=true,breaklinks=true,letterpaper=true,colorlinks,bookmarks=false]{hyperref}

 \cvprfinalcopy 

\def\cvprPaperID{1252} 

\ifcvprfinal\fi
\begin{document}


\title{Track and Transfer: Watching Videos to Simulate Strong\\Human Supervision for Weakly-Supervised Object Detection}

\author{Krishna Kumar Singh, Fanyi Xiao, and Yong Jae Lee\\
University of California, Davis
}

\maketitle
\thispagestyle{empty}

\begin{abstract}
\vspace*{-0.13in}
The status quo approach to training object detectors requires expensive bounding box annotations.  Our framework takes a markedly different direction: we transfer tracked object boxes from weakly-labeled videos to weakly-labeled images to automatically generate \emph{pseudo ground-truth boxes}, which replace manually annotated bounding boxes.  We first mine discriminative regions in the weakly-labeled image collection that frequently/rarely appear in the positive/negative images.  We then match those regions to videos and retrieve the corresponding tracked object boxes.  Finally, we design a hough transform algorithm to vote for the best box to serve as the pseudo GT for each image, and use them to train an object detector.  Together, these lead to state-of-the-art weakly-supervised detection results on the PASCAL 2007 and 2010 datasets.
\vspace*{-0.18in}
\end{abstract}

\section{Introduction}
\label{sec:introduction}
\vspace*{-0.05in}

Object detection is a fundamental problem in computer vision.  While tremendous advances have been made in recent years, existing state-of-the-art methods~\cite{pedro-dpm,overfeat,rcnn,fast-rcnn} are trained in a strongly-supervised fashion, in which the system learns an object category's appearance properties and precise localization information from images annotated with bounding boxes. However, such carefully labeled exemplars are expensive to obtain in the large numbers that are needed to fully represent a category's variability, and methods trained in this manner can suffer from unintentional biases or errors imparted by annotators that hinder the system's ability to generalize to new, unseen data~\cite{dataset-bias}.

\begin{figure}[t]
    \centering
    \includegraphics[width=0.47\textwidth]{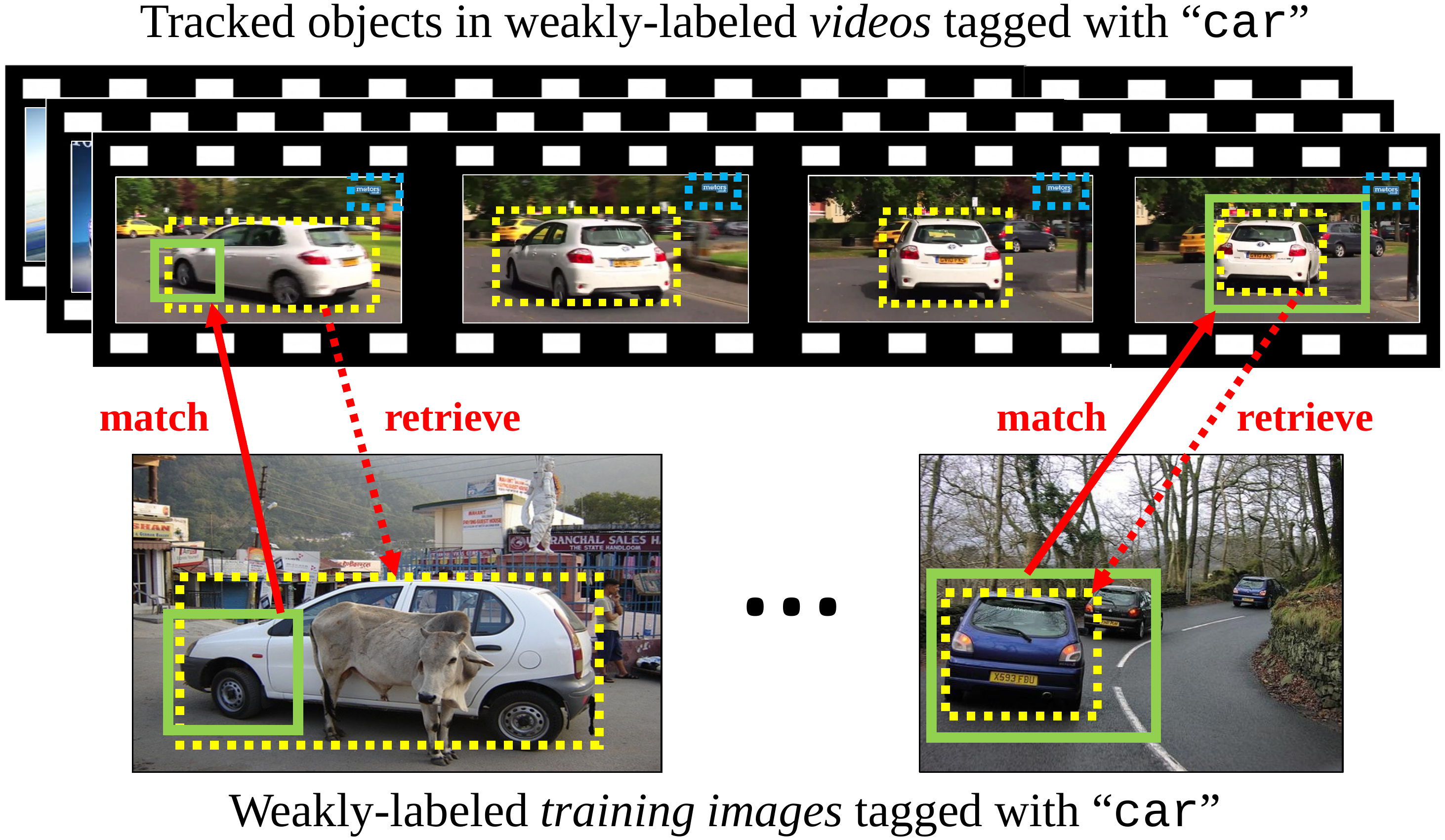}
    \caption{\small{\textbf{Main idea.}  (\textbf{top}) Automatically tracked objects (yellow and blue boxes) in weakly-labeled videos \emph{without any human initialization}.  (\textbf{bottom}) Discriminative visual regions (green boxes) mined in weakly-labeled training images.  For each discriminative region, we find its best matching region across all videos, and retrieve its overlapping tracked object box (yellow dotted box) back to the image.  The retrieved boxes are used as \emph{pseudo ground-truth} to train an object detector.  Our approach improves object localization by expanding the initial visual region beyond a small object part (\textbf{bottom-left}) or removing the surrounding context (\textbf{bottom-right}).  In practice, we combine the retrieved boxes from multiple visual regions in an image to produce its best box.}}
    \label{fig:concept}
    \vspace{-0.1in}
\end{figure}

To address these issues, researchers have proposed to train object detectors with relatively inexpensive \emph{weak supervision}, in which each training image is only weakly-labeled with an image-level tag (e.g., ``car'', ``no car'') that states an object's presence/absence but not its location~\cite{weber-eccv2000,fergus-cvpr2003,pandey-iccv2011,siva-eccv2012,song-icml2014,cinbis-cvpr2014}.  These methods typically mine discriminative visual patterns in the training data that frequently occur in the images that contain the object and rarely in the images that do not.  However, due to scene clutter, intra-class appearance variation, and occlusion, the discriminative patterns often do not tightly fit the object-of-interest; they either correspond to a small part of the object such as a car's wheel instead of the entire car, or include the surrounding context such as a car with portions of the surrounding road (Fig.~\ref{fig:concept} bottom, green boxes).  Consequently, the detector that is trained using these patterns performs substantially worse than strongly-supervised algorithms.

\vspace{-12pt}
\paragraph{Main idea.}  So, how can we create accurate object detectors that do not require expensive bounding box annotations? Our key idea is to use motion cues from videos as a \emph{substitute for strong human supervision}.  Given a weakly-labeled image collection and videos retrieved using the same weak-label (e.g., ``car''), we first automatically track and localize candidate objects in the videos, and then transfer their relevant tracked object boxes to the images.  We transfer the object boxes by mining discriminative visual regions in the image collection, and then matching them to regions in the videos.  See Fig.~\ref{fig:concept}.

Since temporal contiguity and motion signals are leveraged to localize and track the objects in video, their transferred boxes can provide precise object localizations in the weakly-labeled images.  Specifically, they can expand the initial discovered region to provide a fuller coverage of the object, or decrease the spatial extent of the initial discovered region to remove the surrounding context (Fig.~\ref{fig:concept} bottom, yellow boxes).  We then use the transferred boxes to generate \emph{pseudo ground-truth} bounding boxes on the weakly-labeled images to train an object detector, replacing standard human-annotated bounding boxes.  To account for noise in the discovered discriminative visual regions, video tracking, and image-to-video matches, we retrieve a large set of object boxes and combine them with a hough transform algorithm to produce the best boxes.

What is the advantage of transferring object boxes to images instead of directly learning from videos?  In general, images provide more diverse \emph{intra-category} appearance information than videos, especially given the same amount of data (e.g., a 1000-frame video with a single object instance vs.\@ 1000 images with $\sim$1000 different object instances), and are often of higher quality since frames from real-world (e.g., YouTube) videos typically suffer from motion blur and compression artifacts.  Importantly, in this way, our framework opens up the possibility to leverage the huge \emph{static} imagery available online, much of which is already weakly-labeled.

\vspace{-12pt}
\paragraph{Contributions.}  In contrast to existing strongly-supervised object detection systems that require expensive bounding box annotations, or weakly-supervised systems that rely solely on appearance-based grouping cues within the image dataset, we instead transfer tracked object boxes from videos to images to serve as \emph{pseudo ground-truth} to train an object detector.  This eliminates the need for expensive bounding box annotations, and compared to existing weakly-supervised algorithms, our approach provides more complete and tight localizations of the discovered objects in the training data.  Using videos from the YouTube-Objects dataset~\cite{prest-cvpr2012}, we demonstrate that this leads to state-of-the-art weakly-supervised object detection results on the PASCAL VOC 2007 and 2010 datasets.

\vspace*{-0.01in}
\section{Related Work}
\label{sec:relatedwork}
\vspace*{-0.05in}

\paragraph{Weakly-supervised object detection.}  While recent state-of-the-art \emph{strongly-supervised} methods~\cite{rcnn,overfeat,spp-net,fast-rcnn} using deep convolutional neural networks (CNN)~\cite{lecun-1989,krizhevsky-nips2012} have shown great object detection accuracy, they require thousands of expensive bounding-box annotated images.

\begin{figure*}[th!]
    \centering
    \includegraphics[width=1\textwidth]{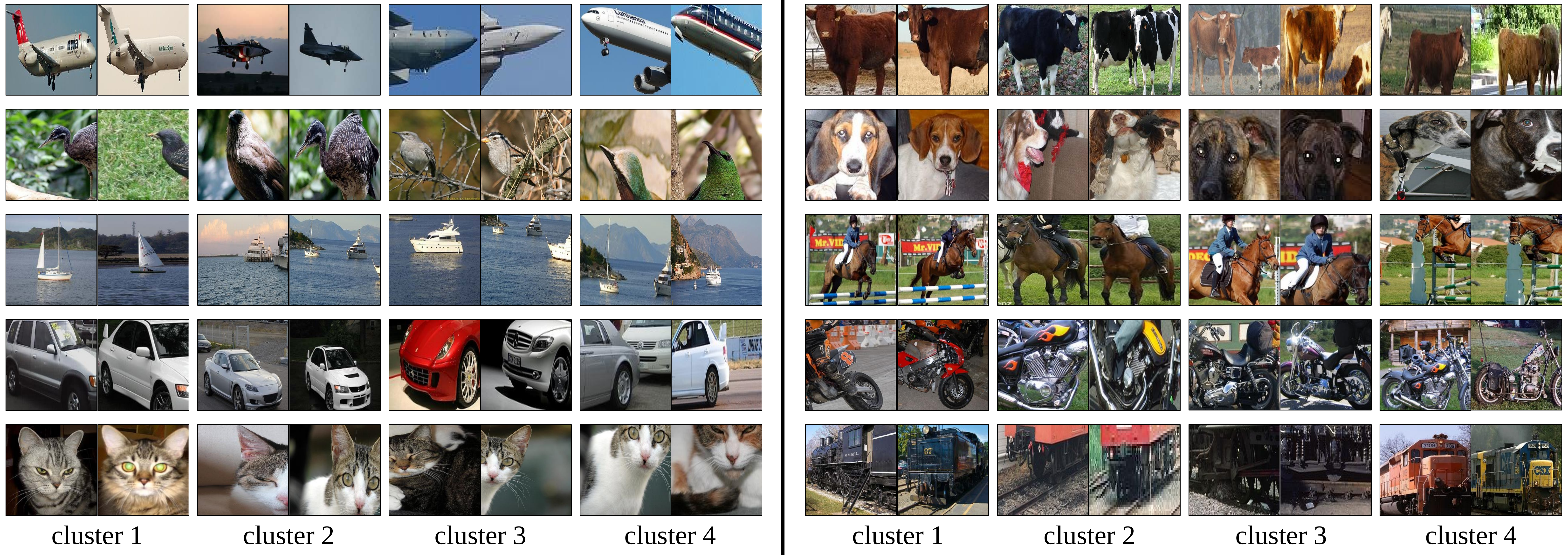}
    \caption{\small{Example positive regions in the top-4 automatically mined discriminative clusters for aeroplane, bird, boat, car, cat, cow, dog, horse, motorbike, and train.  While the discovered regions are relevant to the positively-labeled object category, most of them do not localize the object well, capturing only an object-part (e.g., cat, cluster 1) or including the surrounding context (e.g., aeroplane, cluster 2).}}
    \label{fig:clusters}
    \vspace{-0.1in}
\end{figure*}

To alleviate expensive annotation costs, weakly-supervised methods~\cite{weber-eccv2000,fergus-cvpr2003,pandey-iccv2011,siva-eccv2012,song-icml2014,cinbis-cvpr2014} train models on images labeled only with object presence/absence labels, without any location information of the object.  Early efforts~\cite{weber-eccv2000,fergus-cvpr2003} focused on simple datasets with a single prominent object in each image (e.g., Caltech-101). Since then, a number of methods~\cite{deselaers-eccv2010,pandey-iccv2011,siva-eccv2012,song-icml2014,song-nips2014,cinbis-arxiv2015,oquab-cvpr2015} learn detectors on more realistic and challenging datasets (e.g., PASCAL VOC~\cite{pascal-voc}).  The main idea is to identify discriminative regions that frequently appear in positive images and rarely in negative ones.  However, their central weakness is that due to large intra-category appearance variations, occlusion, and background clutter, they often mis-localize the objects in the training images, which results in sub-optimal detectors.   We address this challenge by matching the discriminative regions to videos to retrieve automatically-tracked object boxes back to the images.  This results in better localization on the weakly-labeled training set, which leads to more accurate object detectors.

\vspace{-12pt}
\paragraph{Learning with videos.}  Video offers something that static images cannot: it provides motion information, a strong cue for grouping objects (the ``law of common fate'' in Gestalt psychology).  Existing methods learn part-based animal models~\cite{ramanan-pami2006}, learn detectors from images while using video patches for regularization~\cite{leistner-cvpr2011}, or augment training data from videos for single-image action recognition~\cite{chen-cvpr2013}.  While some work consider learning object category models directly from (noisy) internet videos~\cite{prest-cvpr2012,hartmann-eccv2012,misra-cvpr2015}, we are exploring a rather different problem: we use video data to simulate human annotations, but ultimately use \emph{image data} to train our models.  Critically, this allows our framework to potentially take advantage of the huge static image data available on the Web, which existing video-only learning methods cannot.

Finally, recent work uses videos for semi-supervised object detection with bounding box annotations as initialization~\cite{liang-arxiv2015}, or trains a CNN for feature learning using tracking as supervision and fine-tuning the learned representation with bounding box annotations for detection~\cite{wang-arxiv2015}.  In contrast, we do not require \emph{any} bounding box annotations.

\vspace*{-0.01in}
\section{Approach}
\label{sec:approach}
\vspace*{-0.05in}

We are given a weakly-labeled image collection $S_I{=}\{I_1,\dots,I_N\}$, in which images that contain the object-of-interest (e.g., ``car'') are labeled as positive and the remaining images are labeled as negative. We are also given a weakly-labeled video collection $S_V{=}\{V_1,\dots,V_M\}$ whose videos contain the positive object-of-interest, but where and when in each video it appears is unknown.

There are three main steps to our approach: (1) identifying discriminative visual regions in $S_I$ that are likely to contain the object-of-interest; (2) matching the discriminative regions to tracked objects in videos in $S_V$ and retrieving the tracked objects' boxes back to the images in $S_I$; and (3) training a detector using the images in $S_I$ with the retrieved object boxes as supervision.  

\subsection{Mining discriminative positive visual regions}
\label{subsec:mining}
\vspace{-0.05in}

We first mine discriminative visual regions in the image collection $S_I$ that frequently appear in the positive images and rarely in the negative ones; these regions will likely correspond to the object-of-interest or a part of it.  For this, we follow a similar approach to~\cite{singh-eccv2012,doersch-siggraph2012,song-icml2014,song-nips2014}.

For each image in $S_I$, we generate $\sim$2000 object proposals (rectangular regions) using selective search~\cite{selectivesearch}, and describe each proposal with a \emph{pool5} activation feature using AlexNet~\cite{krizhevsky-nips2012} pre-trained for ImageNet classification.  For each region, we find its best matching (nearest neighbor) region in each image in $S_I$ (regardless of image label) using cosine similarity.  Each region and its $k$ closest nearest neighbors form a cluster.  We then rank the clusters in descending order of the number of cluster instances that are from the positive images.  Since we create clusters for every region in every image, many will be redundant.  We therefore greedily remove near-duplicate clusters that contain many near-identical regions to any higher-ranked cluster, as measured by spatial overlap of more than 25\% IOU between 10\% of their cluster members.  Finally, for each remaining cluster, we discard any negative regions.

Let $\mathcal{P}$ be the set of all positive regions in the top-$C$ ranked clusters.  While $\mathcal{P}$ contains many diverse and discriminative regions of the object-of-interest (see Fig.~\ref{fig:clusters}), most of the regions will not tightly \emph{localize} the object for three main reasons: (1) the most discriminative regions usually correspond to object-parts, which tend to have less appearance variation than the full-object (e.g., face vs.~full-body of a cat), (2) co-occurring ``background'' objects are often included in the region (e.g., airplane with sky), and (3) most of the initial object proposals are noisy and do not tightly fit any object to begin with.  Thus, the regions in $\mathcal{P}$ will be sub-optimal for training an object detector, since they are not well-localized; this is the central weakness of all existing weakly-supervised methods.  We next explain how to use videos labeled with the same weak-label (e.g., ``car'') to improve the localization.

\subsection{Transferring tracked object boxes}
\label{subsec:transfer}
\vspace{-0.05in}

For now, assume that we have a (noisy) object track in each video in $S_V$, which fits a bounding box around the positive object in each frame that it appears.  In Sec.~\ref{subsec:tracking}, we explain how to obtain these tracks.

\begin{figure*}[t!]
    \centering
    \includegraphics[width=1\textwidth]{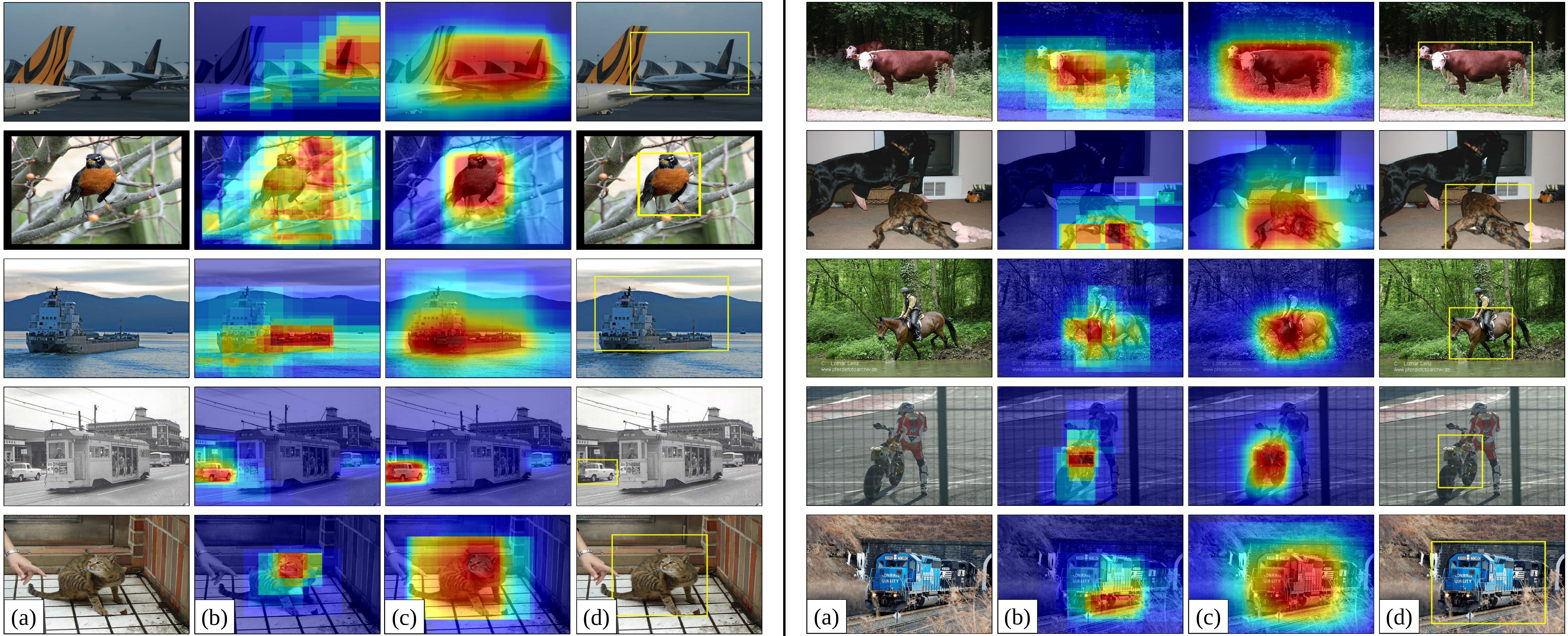}
    \caption{\small{(a) Weakly-labeled positive image for aeroplane, bird, boat, car, cat, cow, dog, horse, motorbike, and train.  (b) Heatmap showing the distribution of the initial discriminative positive regions found in the image.  (c) Heatmap showing the distribution of the transferred video object boxes in the image.  (d) Our automatically discovered pseudo ground-truth box.  Notice how the initial discriminative regions focus more on object-parts, whereas the transferred boxes focus more on the full object.  This leads to better localization of the object in the weakly-labeled positive image.  \textbf{Best viewed on pdf.}  \emph{Results for more images can be found at the end of the paper.}}}
    \label{fig:bbtransfer}
    \vspace{-0.1in}
\end{figure*}

For each positive image region in $\mathcal{P}$, we search for its $n$ best matching video regions across all videos in $S_V$ and return their corresponding tracked object boxes to improve the localization of the object in its image.  There is an important detail we must address to make this practical: matching with \emph{fc7} features (of AlexNet~\cite{krizhevsky-nips2012}) can be prohibitively expensive, since each candidate video region (e.g., selective search proposal) would need to be warped to 227x227 and propagated through the deep network, and there can be $\sim$2000 such candidate regions in every frame, and millions of frames.  Instead, we perform matching with \emph{conv5} features, which allows us to forward-propagate an entire video frame just once through the network since convolutional layers do not require fixed-size inputs.  To compute the \emph{conv5} feature maps, we use deep pyramid~\cite{deeppyramid}, which creates an image pyramid with 7 levels (where the scale factor between levels is $2^{-1/2}$) and computes a \emph{conv5} feature map for each level (for the 1st level, the input frame is resized such that its largest dimension is 1713 pixels).  We then match each positive image region to each frame in each video densely across location and scale in a sliding-window fashion in \emph{conv5} feature space, using cosine similarity.  Note that this restricts matching between regions with similar aspect ratios, which can also help reduce false positive matches.

\begin{figure}[t!]
    \centering
    \includegraphics[width=0.47\textwidth]{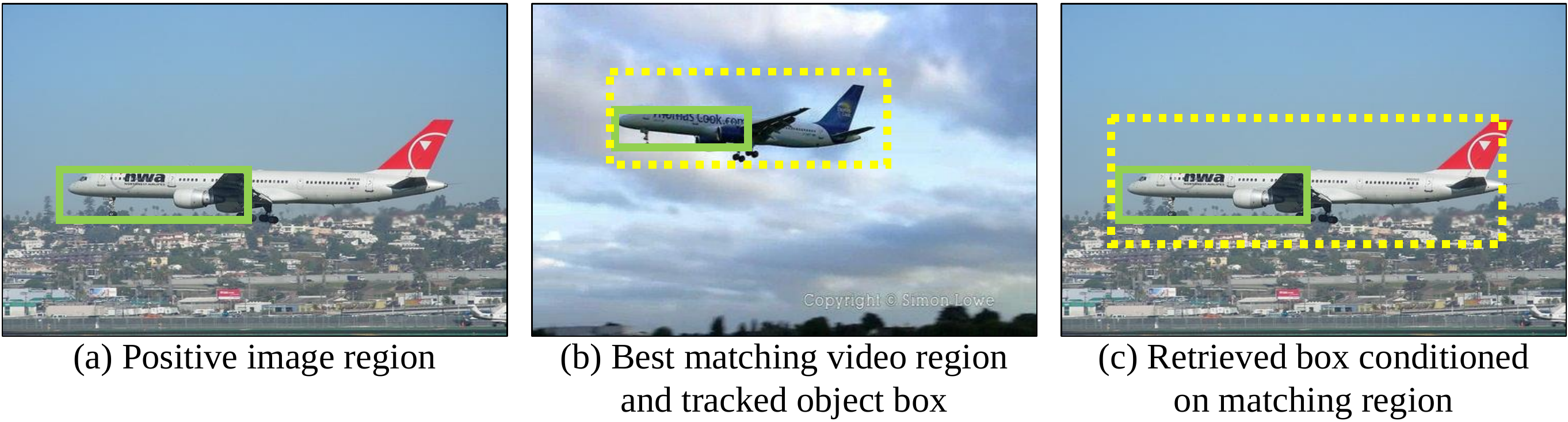}
    \caption{\small{We match a positive image region (a) to all video frames in a sliding-window fashion, and for the best matching vido region (green box) (b), we retrieve its overlapping tracked object box (yellow dotted box) back to the image (c).}}
    \label{fig:video_match}
    \vspace{-0.15in}
\end{figure}

Given a positive image region's $n$ best matching video regions, we return each of their frame's tracked object bounding box (if it has any spatial overlap with the matched video region) back to the positive region's image, while preserving relative translation and scale differences.  Specifically, we can parameterize any region with its top-left and bottom-right coordinate values: $[x_{min},y_{min},x_{max},y_{max}]$.  Denote a positive image region as $\vct{r}$, its matched video region as $\vct{v}$, and the corresponding overlapping tracked region as $\vct{t}$.  Then, the returned bounding box $\vct{r'}$ is: $\vct{r'} = \vct{r} + (\vct{t} - \vct{v})$.  See Fig.~\ref{fig:video_match}.  We repeat this for all $n$ best matching video regions, and for each positive region in $\mathcal{P}$.

Each positively-labeled image in $S_I$ (that has at least one positive region) now has a set of retrieved bounding boxes, up to $n$ from each positive region in the image.  Some will tightly fit the object-of-interest, while others will be noisy due to incorrect matches/tracks.  We thus use the hough transform to vote for the best box in each image.  Specifically, we create a 4-dimensional hough space in which each box casts a vote for its $[x_{min},y_{min},x_{max},y_{max}]$ coordinates.  We select high density regions in the continuous hough space with mean-shift clustering~\cite{meanshift}, which helps the voting be robust to noise and quantization errors~\cite{leibe-ijcv2008}.  The total vote for box coordinate $\vct{l}$ is a weighted sum of the votes in its spatial vicinity:
\vspace{-0.02in}
\begin{equation}
vote(\vct{l}) = \sum_i vote(\vct{r'}_i) \cdot K\Big(\frac{\vct{l}-\vct{r'}_i}{b}\Big),
\vspace{-0.02in}
\end{equation}
where the kernel $K$ is a radially symmetric, non-negative function centered at zero and integrating to one, $b$ is the mean-shift kernel bandwidth, $i$ indexes over the positive regions in the image, and $vote(\vct{r'}_i)$$=$$1, \forall i$.  We select $\vct{l}$ with the highest vote as the final box for the image.  If the highest vote is less than a threshold $\theta=20$, then there is not enough evidence to trust the box so we discard it.  See Fig.~\ref{fig:bbtransfer} (c-d) for example distributions of the transferred bounding boxes and final selected bounding box.  We repeat this hough voting process for each positively-labeled image in $S_I$.  

\subsection{Training an object detector}
\label{subsec:training}
\vspace{-0.05in}

We can now treat the final selected boxes as pseudo ground-truth (GT)---as a \emph{substitute for manually annotated boxes}---to train an object detector, with any algorithm developed for the strongly-supervised setting.  We use the state-of-the-art Regions with CNN (R-CNN) system~\cite{rcnn}.  Briefly, R-CNN computes CNN features over selective search~\cite{selectivesearch} proposals, trains a one-vs-all linear SVM (with GT boxes as positives and proposals that have less than 0.3 intersection-over-union overlap (IOU) with any GT box as negatives) to classify each region, and then performs bounding box regression to refine the object's detected location.

There are three considerations to make when adapting R-CNN to our work: (1) each positively-labeled image has at most one pseudo GT box, which means that negative regions from the same image must be carefully selected since the image could have multiple positive instances (e.g., multiple cars in a street scene) but our pseudo GT may only be covering one of them; (2) some positively-labeled images may have no pseudo GT box (i.e., if there were not enough votes), which means that we would not be making full use of all the positive images; and (3) some pseudo GT boxes may be inaccurate even after hough voting due to noise in the matching or tracking.  These can all lead to a sub-optimal detector if not handled carefully.

To address the first issue, we train an R-CNN model with the pseudo GT boxes as positives, and any selective search proposal that has an IOU less than 0.3 \emph{and} greater than 0.1 with a pseudo GT box as negatives.  In this way, we minimize the chance of mistakenly labeling a different positive instance in the image as negative, but at the same time, select mis-localized regions (that have some overlap with a pseudo GT) as \emph{hard-negatives}.  We treat all selective search proposals in any negatively-labeled image in $S_I$ as negative.

To address the second and third issues, we perform a latent SVM (LSVM) update~\cite{pedro-dpm} given the initial R-CNN model from above to update the pseudo GT boxes.  For images that do not have a pseudo GT box, we fire the R-CNN model and take its highest-scoring detection in the image as the pseudo GT box.  For images that already have a pseudo GT box, we take the highest-scoring detection that has at least 0.5 IOU with it, which prevents the updated box from changing too much from the initial box.  We then re-train the R-CNN model with the updated pseudo GT boxes.

Finally, we also fine-tune the R-CNN model to update not only the classifier but also the features using our pseudo GT boxes, which results in an even greater boost in detection accuracy (as shown in Sec.~\ref{subsec:detectionaccuracy}).  Fine-tuning CNN features has not previously been demonstrated in the weakly-supervised detection setting, likely due to existing methods producing too many false detections in the training data.  Our discovered pseudo GT boxes are often quite accurate, making our approach amenable for fine-tuning.

\begin{table*}[t!]
\begin{center}
    \footnotesize
    \begin{tabular}{| c | c c c c c c c c c c | c |}
    \hline    	
    \textbf{VOC 2007 train+val} & aero & bird & boat & car & cat & cow & dog & horse & mbike & train & mean CorLoc \\
    \hline
    Initial pseudo GT (with all images) &  48.8 & 33.9 & 13.3 & 57.3 & 46.5 & 32.2 & 44.4 & 40.8 & 48.2 & 43.7 & 40.9 \\
    Initial pseudo GT (excluding missed images) & 58.8 & 49.6 & 17.7 & 64.7 & 60.4 & 44.8 & 52.8 & 55.3 & 54.3 & 53.0 &  51.1 \\
    Updated pseudo GT (with all images) & 58.8 & 49.6 & 15.4 & 64.9 & 59.0 & 43.2 & 51.2 & 57.5 & 63.1 & 54.4 & 51.7 \\
    \hline
    \end{tabular}
    \caption{Localization accuracy in terms of CorLoc on the VOC 2007 train+val set.  We evaluate our initial and updated pseduo GT boxes.  The final boxes (third row) provide very good localizations in the training data, which leads to accurate training of object detectors.}
    \label{table:corloc}
\end{center}
\vspace*{-0.2in}
\end{table*}

\subsection{Unsupervised video object tracking}
\label{subsec:tracking}
\vspace{-0.05in}

Our framework requires an accurate unsupervised video object tracker, since its tracked object boxes will be used to generate the pseudo GT boxes on the weakly-labeled images.  For this, we use the unsupervised tracking method of~\cite{fanyitube}, which creates a diverse and representative set of spatial-temporal object proposals in an unannotated video.  Each spatial-temporal proposal is a sequence of boxes fitting an object over multiple frames in time.\footnote{We also tried the video segmentation method of~\cite{ochs-pami2014}.  However, it fails to produce good segmentations when an object is not moving.  Ultimately, transferring its object boxes resulted in a slightly worse detector.}

Briefly, the method begins by leveraging appearance and motion objectness to score a set of static object proposals in each frame, and then groups high-scoring proposals across frames that are similar in appearance and frequently appear throughout the video.  Each group is then ranked according to the average objectness score of its instances.  For each group, the method trains a discriminative tracking model with the group's instances as positives and all non-overlapping regions in their frames as negatives, and tracks the object in each instance's adjacent frames.  The model is then retrained with the newly tracked instances as positives, and the process iterates until all frames are covered.  The output is a set of ranked spatio-temporal tracks that fit a box around the objects in each frame that they appear.  The method also has a pixel-segmentation refinement step, but we skip it for speed.  See~\cite{fanyitube} for details.

\begin{figure}[t!]
    \centering
    \includegraphics[width=0.47\textwidth]{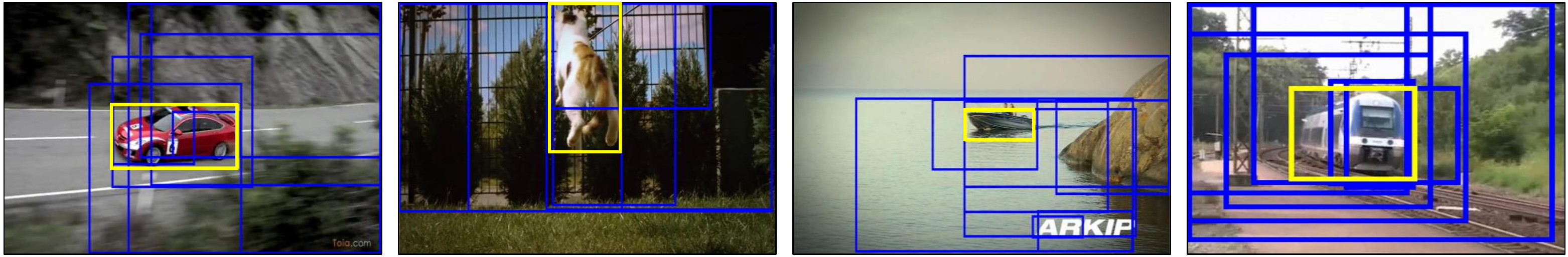}
    \caption{\small{Examples showing the spatio-temporal boxes generated with~\cite{fanyitube} (blue), and our automatically selected box (yellow).}}
    \label{fig:fanyi_tubes}
    \vspace{-0.1in}
\end{figure}

For each video in $S_V$, we take the 9 highest-ranked tracks generated by~\cite{fanyitube}.  Not all of these tracks will correspond to the object-of-interest.  We therefore use our mined positive regions in $\mathcal{P}$ to try to select the relevant one in each frame.  Specifically, given frame $f$, we match each positive region $\vct{r}_i$ to it in a sliding-window fashion in \emph{conv5} feature space (as in Sec.~\ref{subsec:transfer}), and record its best matching box $\vct{v}_i^f$ in the frame.  We score a tracked box $\vct{t}_j^f$ in frame $f$ as: $score(\vct{t}_j^f) = \sum_i \mbox{IOU}(\vct{v}_i^f,\vct{t}_j^f) \times sim(\vct{r}_i,\vct{v}_i^f)$, where $i$ indexes the positive regions in $\mathcal{P}$, $j$ is the index of a tracked video box, and $sim$ is cosine similarity.  We choose the tracked box with the highest score, and discard the rest.  Our selection criterion favors choosing a box in each video frame that has high-overlap with many good matches from discriminative positive regions. See Fig.~\ref{fig:fanyi_tubes} for examples.  The selected video boxes are provided as input to the video-matching module described in Sec.~\ref{subsec:transfer}.

\section{Experiments}
\vspace*{-0.05in}

We analyze: (1) localization accuracy of our discovered pseudo GT boxes on the weakly-labeled training images, (2) detection performance of our trained models on the test images, (3) ablation studies analyzing the different components of our approach, and (4) our selection criterion for choosing the relevant object track in each video frame.

\vspace*{-12pt}
\paragraph{Datasets.}  We use videos from YouTube-Objects~\cite{prest-cvpr2012} and images from PASCAL VOC 2007 and 2010.  We evaluate on their \textbf{10 shared classes} (treating each as a positive in turn): aeroplane, bird, boat, car, cat, cow, dog, horse, motorbike, train.  YouTube-Objects contains 9-24 videos per class; each video is 30-180 sec; 570K total frames. We only use each video's weak category-label (i.e., we do not know in which frames or regions the object appears). Each video is divided into shots with similar color~\cite{prest-cvpr2012}; we generate object tracks for each shot using~\cite{fanyitube}.  VOC 2007 is used by all existing state-of-the-art weakly-supervised detection algorithms; VOC 2010 is used by~\cite{cinbis-arxiv2015}.  For VOC 2007 and 2010, we use the train+val (5011 imgs) and train set (4998 imgs), respectively, to discover the pseudo GT boxes.  For both datasets, we report detection results on the test set using average precision.  In contrast to existing weakly-supervised methods (except~\cite{song-icml2014,song-nips2014}), we do not discard instances labeled as \emph{pose, difficult, truncated}, and restrict the supervision to the image-level object presence/absence labels to mimic a more realistic (difficult) weakly-supervised scenario.

\vspace*{-12pt}
\paragraph{Implementation details.}
For mining discriminative regions, we take $k$$=$$(\# \mbox{ positive images})/2$ nearest neighbors, and top $C$$=$200 clusters.  When matching a positive region to video, we adjust its box to have roughly 48 \emph{conv5} cells using a sizing heuristic~\cite{exemplar-svm}, and compute matches in every 8th frame for speed.  For the mean-shift bandwidth $b$, we train separate detection models for $b$$=$$[100,250,500,1000]$ and validate detection accuracy over our automatically selected object tracks on YouTube-Objects (i.e., we treat them as noisy GT); even though the discovered tracks can be noisy, we find they produce sufficiently good results for cross-validation.  To compute deep features, we use AlexNet pre-trained on ILSVRC 2012 classification, using Caffe~\cite{krizhevsky-nips2012,jia-arxiv2014}.  We \emph{do not} use the R-CNN network fine-tuned on PASCAL data~\cite{rcnn}.

To fine-tune our detector, we take our discovered pseudo GT boxes over all 10 categories to fine-tune the CNN (AlexNet pre-trained on ILSVRC2012 classification) by replacing its 1000-way classification layer with a randomly-initialized 11-way classification layer (10 categories plus background).  We treat all selective search proposals with 0.6$\ge$IOU with a pseudo GT box as positives for that box's category, and all proposals with 0.1$\le$IOU$\le$0.3 with a pseudo GT box as negatives.  All proposals from images not belonging to any of the 10 categories are also treated as negatives.  We start SGD at a learning rate of 0.001 and decrease by $\times\frac{1}{10}$ after 20,000 iterations. In each SGD iteration, 32 positives (over all classes) and 96 negatives are uniformly sampled to construct a mini-batch. We perform 40,000 SGD iterations.

\begin{figure*}[t!]
    \centering
    \includegraphics[width=1\textwidth]{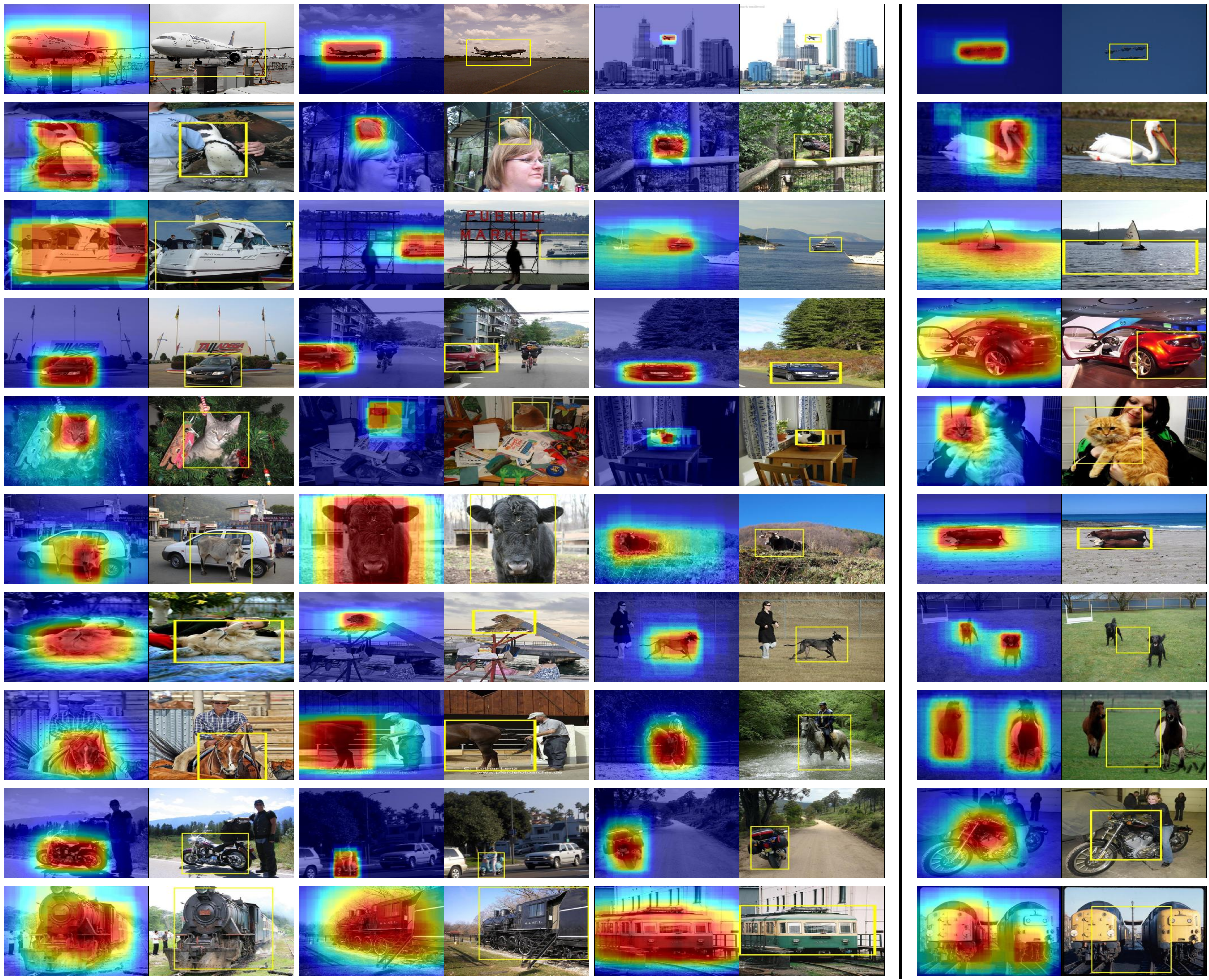}
    \caption{\small{Qualitative results on the VOC 2007 train+val set.  In each image pair, the first image shows a heatmap of the transferred video object boxes and the second image shows the final selected pseudo ground-truth box.  Our approach accurately discovers the spatial extent of the object-of-interest in most of the images.  The last column shows mis-localized examples.  Our approach can fail when there are multiple instances of the same object category in the image (e.g., aeroplane, dog, horse, train) or when the object's appearance is very different from that found in videos (e.g., car).  \textbf{Best viewed on pdf.}}}
    \label{fig:pseudo_gt_results}
    \vspace{-0.1in}
\end{figure*}

\subsection{Pseudo ground-truth localization accuracy}
\label{subsec:pseudogtaccuracy}
\vspace{-0.05in}

We first analyze the localization accuracy of our discovered pseudo GT boxes on the VOC 2007 train+val dataset.  We use the correct localization (CorLoc) measure~\cite{deselaers-eccv2010}, which is the fraction of positive training images in which the predicted object box has an intersection-over-union overlap (IOU) greater than 50\% with any ground-truth box.  As mentioned in~\cite{cinbis-arxiv2015}, CorLoc is not consistently measured across previous studies, due to changes in the training sets (for example, we do not exclude the images annotated as \emph{pose, difficult, truncated}).  Thus, we only use it to analyze our own pseudo GT boxes, and use detection accuracy to compare against the state-of-the-art.

Table~\ref{table:corloc} shows the results.  Our initial pseudo GT boxes produce an average CorLoc score of 40.9\% across all categories (first row).  However, we initially miss discovering a pseudo GT box in 12\% of the images, which pulls down the average. (Recall we only keep the most confident box in each image that has at least $\theta=20$ votes.)  If we only consider the images in which a pseudo GT is initially found, then our average increases to 51.1\% (second row).  By detecting the missed pseudo GT boxes and updating the existing ones using the R-CNN model trained with the initial pseudo GT boxes (via an LSVM update), our final CorLoc average improves to 51.7\% (third row).  For the boat category, our low performance is due to boats often occurring with water; since water seldom appears in other categories, many water regions are mistakenly found to be discriminative, which leads to inaccurate localizations of the boat (see Sec.~\ref{subsec:locerroranalysis} for a further detailed breakdown of the error cases per class).  For the remaining categories, our pseudo GT boxes localize the objects well, and we will see in Sec.~\ref{subsec:detectionaccuracy} that they lead to robust object detectors.

\subsection{Pseudo ground-truth visualization}
\label{subsec:visualization}
\vspace{-0.05in}

We next visualize our discovered pseudo GT on the VOC 2007 train+val set.  In each image pair in Fig.~\ref{fig:pseudo_gt_results}, we display a heatmap of the transferred video object boxes and the final selected pseudo GT box.  Our method accurately localizes the object-of-interest in many images, even in difficult cases where the object is in an atypical pose (1st dog), partially-occluded (2nd car), or in a highly-cluttered scene (2nd cat).  The last column shows some failure cases.  The most prominent failure case is when there are multiple instances of the same object category that are spatially close to each other.  This is due to a sub-optimal mean-shift bandwidth parameter $b$, which is used in the voting of the pseudo GT box. Although we automatically select $b$ via cross-validation on the video tracks (see implementation details), it is fixed \emph{per-category}.  Using an adaptive bandwidth~\cite{comaniciu-iccv2001} to automatically find an optimal value \emph{per-image} may help to alleviate such errors.  Importantly, these errors occur in only a few images. See end of this paper for more results on randomly chosen images per class.

Overall, the qualitative results demonstrate that by transferring object boxes from automatically tracked objects in video, we can accurately discover the objects' full spatial extent in the weakly-labeled image collection.

\subsection{Weakly-supervised detection accuracy}
\label{subsec:detectionaccuracy}
\vspace{-0.05in}

We next compute detection accuracy using the R-CNN model trained using our pseudo GT boxes.  We compare with state-of-the-art weakly-supervised detection methods~\cite{song-icml2014,song-nips2014,bilen-bmcv2014,wang-eccv2014,cinbis-arxiv2015} that use the same AlexNet CNN features pre-trained on ILSVRC 2012. Note that our approach and the previous methods all use the same PASCAL VOC training images to train the detectors.  Our use of videos is only to get better pseudo GT boxes on the training images.

\begin{table*}[t!]
  \begin{minipage}[t]{0.2\textwidth}
    \caption{Detection average precision on the VOC 2007 test set.  We compare our approach to state-of-the-art weakly-supervised methods.}
    \label{table:voc2007}
  \end{minipage}
  \hspace*{0.1in}
  \begin{minipage}[t]{0.67\textwidth}
      \begin{center}
            \footnotesize
            \begin{tabular}{| c | c c c c c c c c c c | c |}
            \hline    	
            \textbf{VOC 2007 test} & aero & bird & boat & car & cat & cow & dog & horse & mbike & train & mAP \\
            \hline
            Song et al., 2014~\cite{song-icml2014} & 27.6 & 19.7 & 9.1 & 39.1 & 33.6 & 20.9 & 27.7 & 29.4 & 39.2 & 35.6 & 28.2 \\
            Song et al., 2014~\cite{song-nips2014} & 36.3 & 23.3 & 12.3 & 46.6 & 25.4 & 23.5 & 23.5 & 27.9 & 40.9 & 37.7 & 29.7 \\
            Bilen et al., 2014~\cite{bilen-bmcv2014} & 42.2 & 23.1 & 9.2 & 45.1 & 24.9 & 24.0 & 18.6 & 31.6 & 43.6 & 35.9 & 29.8 \\
            Wang et al., 2014~\cite{wang-eccv2014} & 48.9 & 26.1 & 11.3 & 40.9 & 34.7 & \textbf{34.7} & 34.4 & 35.4 & \textbf{52.7} & 34.8 & 35.4 \\
            Cinbis et al., 2015~\cite{cinbis-arxiv2015} &  39.3 & 28.8 & \textbf{20.4} & 47.9 & 22.1 & 33.5 & 29.2 & 38.5 & 47.9 & 41.0 & 34.9 \\
            Ours w/o fine-tune & 50.7 & 36.6 & 13.4 & 53.1 & 50.8 & 21.6 & 37.6 & 44.0 & 46.1 & 43.4 & 39.7 \\
            Ours & \textbf{53.9} & \textbf{37.7} & 13.7 & \textbf{56.6} & \textbf{51.3} & 24.0 & \textbf{38.5} & \textbf{47.9} & 47.0 & \textbf{48.4} & \textbf{41.9} \\
            \hline
            \end{tabular}
      \end{center}
  \end{minipage}\hfill
  \vspace*{-0.1in}
\end{table*}

\begin{table*}[t!]
  \begin{minipage}[t]{0.2\textwidth}
    \caption{Detection average precision on the VOC 2010 test set.}
    \label{table:voc2010}
  \end{minipage}
  \hspace*{0.1in}
  \begin{minipage}[t]{0.67\textwidth}
      \begin{center}
            \footnotesize
            \begin{tabular}{| c | c c c c c c c c c c | c |}
            \hline    	
            \textbf{VOC 2010 test} & aero & bird & boat & car & cat & cow & dog & horse & mbike & train & mAP \\
            \hline
            Cinbis et al., 2015~\cite{cinbis-arxiv2015} & 44.6 & 25.5 & \textbf{14.1} & 36.3 & 23.2 & 26.1 & 29.2 & 36.0 & \textbf{54.3} & 31.2 & 32.1 \\
            Ours w/o fine-tune & 50.9 & 35.8 & 8.1 & 40.5 & 45.9 & 26.0 & 36.4 & 39.0 & 45.7 & 39.4 & 36.8\\
            Ours & \textbf{53.5} & \textbf{37.5} &  8.0 & \textbf{44.2} & \textbf{49.4} & \textbf{33.7} & \textbf{43.8} & \textbf{42.5} & 47.6 & \textbf{40.6} & \textbf{40.1} \\
            \hline
            \end{tabular}
      \end{center}
  \end{minipage}\hfill
  \vspace*{-0.1in}
\end{table*}

\begin{table*}[t!]
\begin{center}
    \footnotesize
    \begin{tabular}{| c | c c c c c c c c c c | c |}
    \hline    	
    \textbf{VOC 2007 test} & aero & bird & boat & car & cat & cow & dog & horse & mbike & train & mAP \\
    \hline
    Inital pseudo GT & 43.4 & 30.5 & 11.9 & 50.2 & 39.6 & 16.7 & 31.6 & 36.7 & 42.2 & 40.7 & 34.4 \\
    Updated pseudo GT & 48.0 & 34.2 & 12.2 & 51.3 & 43 & 21.9 & 33.4 & 39.1 & 43.8 & 42.2 & 36.9 \\
    Updated pseudo GT + bbox-reg & 50.7 & 36.6 & 13.4 & 53.1 & 50.8 & 21.6 & 37.6 & 44.0 & 46.1 & 43.4 & 39.7 \\
    Updated pseudo GT + fine-tune + bbox-reg & \textbf{53.9} & \textbf{37.7} & \textbf{13.7} & \textbf{56.6} & \textbf{51.3} & \textbf{24.0} & \textbf{38.5} & \textbf{47.9} & \textbf{47.0} & \textbf{48.4} & \textbf{41.9}  \\
    \hline
    \end{tabular}
    \caption{Detection average precision on the VOC 2007 test set to evaluate the different components of our approach.  See text for details.}
    \label{table:ablation}
\end{center}
\vspace*{-0.1in}
\end{table*}

Tables~\ref{table:voc2007} and~\ref{table:voc2010} show results on the VOC 2007 and 2010 test sets, respectively.  Our approach produces the best results with a mAP of 41.9\% and 40.1\%, respectively.  The baselines all share the same high-level idea of mining discriminative patterns that frequently/rarely appear in the positive/negative images.  In particular, the detection results produced by~\cite{song-icml2014} is similar to what we would get if we were to train a detector directly on our initially-mined discriminative positive regions.  Since those regions often correspond to an object-part (e.g., car wheel) or include surrounding context (e.g., car with road) (recall Fig.~\ref{fig:clusters}), these methods have difficulty producing good localizations on the training data, which in turn degrades detection performance.  While~\cite{song-nips2014} tries to combine pairs of discriminative regions to provide better spatial coverage of the object, it is still limited by the mis-localization error of each individual region.  We instead transfer automatically tracked object boxes from weakly-labeled videos to images, which produces more accurate localizations on the training data and leads to higher detection performance.  Our low detection accuracy on cow can be explained by the poor video tracks produced by~\cite{fanyitube} (see Sec.~\ref{subsec:videoaccuracy}), which confirms the need for good object tracks.

Overall, our results suggest a scalable application for object detection, since we can greatly reduce human annotation costs and still obtain reliable detection models.

\begin{table*}[t!]
	\begin{center}
		\footnotesize
		\begin{tabular}{| c | c c c c c c c c c c | c |}
			\hline    	
			\textbf{YouTube-Objects} & aero & bird & boat & car & cat & cow & dog & horse & mbike & train & mean IOU \\
			\hline
			Upper-bound & 67.1 & 70.4 & 56.8 & 77.9 & 63.1 & 23.3 & 64.9 & 67.0 & 65.4 & 62.9 & 61.9 \\
			Our selection scheme & 54.5 & 49.3 & 31.8 & 68.6 & 44.0 & 11.0 & 45.0 & 52.3 & 58.2 & 36.4 & 45.1 \\
			\hline
		\end{tabular}
		\caption{Mean intersection-over-union (IOU) scores of the video object proposals algorithm~\cite{fanyitube} on the YouTube-Objects dataset.  We evaluate the upper-bound performance, and the proposal selected using our selection criterion.}
		\label{table:video}
	\end{center}
	\vspace*{-0.1in}
\end{table*}

\subsection{Ablation studies}
\label{subsec:ablation}
\vspace{-0.05in}

In this section, we conduct ablation studies to tease apart the contribution of each component of our algorithm.  Table~\ref{table:ablation} shows the results.  The first and second rows show mAP detection accuracy produced by the R-CNN models trained using the initial and updated (via LSVM update) pseudo GT boxes, respectively.  The initial R-CNN model produces 34.4\% mAP\@.  Retraining the model with the updated pseudo GT boxes leads to 36.9\% mAP, which shows that the extra positive instances and corrected instances are helpful.  The third row shows bounding box regression results, which further boosts performance to 39.7\% mAP\@.  This confirms that our pseudo GT boxes are well-localized, since the trained bounding box regressor~\cite{pedro-dpm,rcnn} is able to adjust the initial detections to better localize the object.

The last row shows fine-tuning results.  Training an R-CNN model with our fine-tuned features improves results on \emph{all 10 categories} to 41.9\% mAP for VOC 2007.  The improvement is not as significant as in the fully-supervised case, which resulted in a $\sim$9\% point increase for VOC 2007 (see Table~2 in~\cite{rcnn}).  Since our pseudo GT boxes are not perfect, any noise seems to have a more prominent effect than in the fully-supervised case, which has perfect GT boxes.  Still, this result confirms our discovered pseudo GT boxes are quite accurate, making our approach amenable for fine-tuning.


\begin{figure}[t!]
	\centering
	\includegraphics[width=0.47\textwidth]{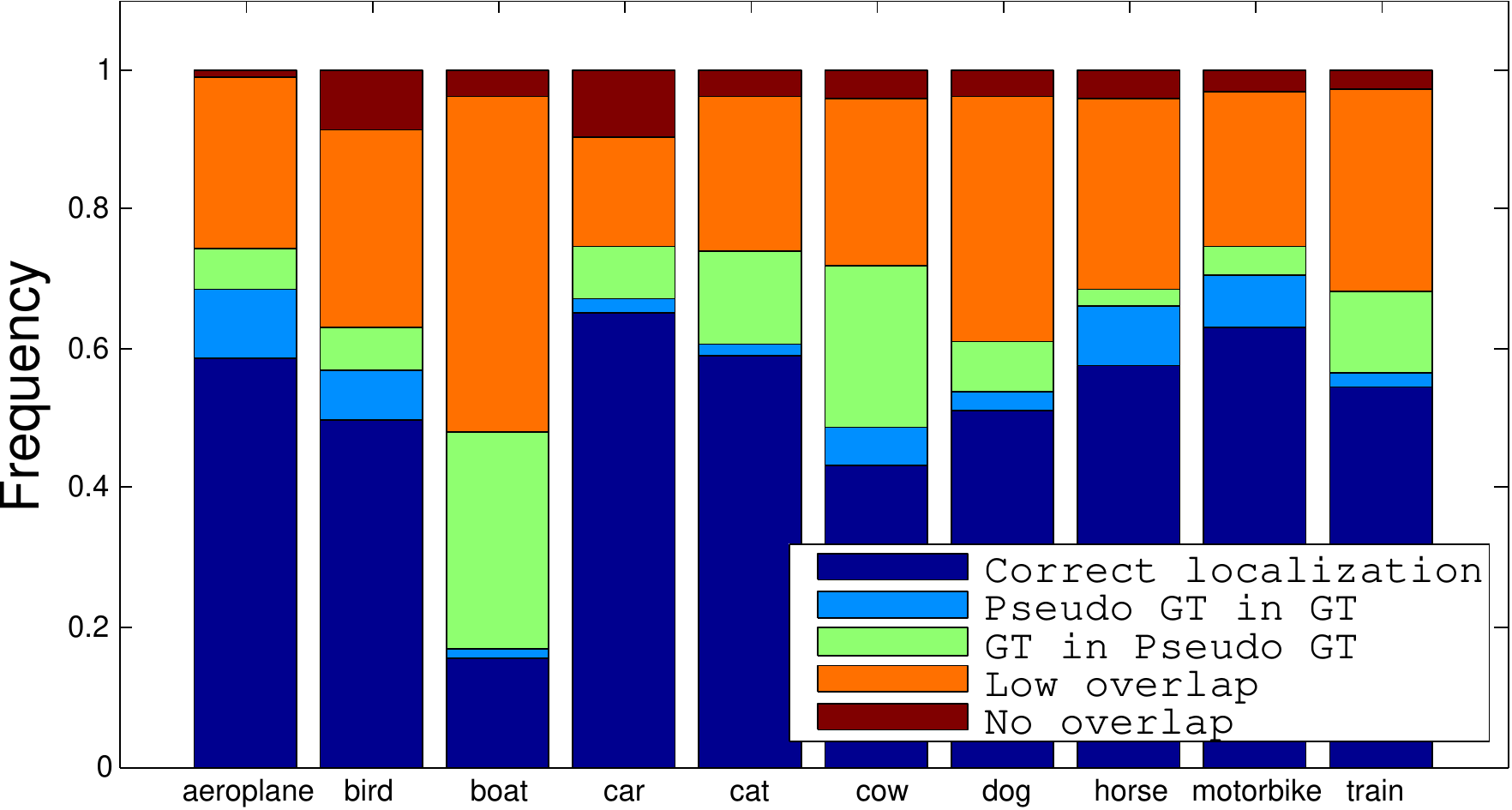}
	\caption{\small{Per category frequency of correct localizations and different localization errors.}}
	\label{fig:error_type}
	\vspace{-0.1in}
\end{figure}

\subsection{Per-category pseudo ground-truth localization error analysis}
\label{subsec:locerroranalysis}
\vspace{-0.04in}

We next expand upon our pseudo GT accuracy analysis (Sec.~\ref{subsec:pseudogtaccuracy}), in order to better understand the localization errors for each category. Following \cite{cinbis-arxiv2015}, we categorize each of our discovered pseudo GT boxes in the positive training images into one of five cases: (i) correct localization (IOU $\geq$ 0.50 with a manually-annotated ground-truth box), (ii) pseudo GT completely inside a ground-truth box, (iii) a ground-truth box completely inside the pseudo GT, (iv) none of the above, but IOU $>$ 0 with a ground-truth box, and (v) no overlap with a ground-truth box (IOU $=$ 0).

Fig.~\ref{fig:error_type} shows the frequency of the five cases for each category. For all categories except boat, our approach is able to correctly localize the object-of-interest in many of the training images (i.e., high frequency of ``€œcorrect localization''€). For boat, the most discriminative regions happen to be water, since it is unique to boat images and does not frequently appear in the other categories. Due to this, our discovered pseudo GT tends to include large portions of water with a boat in it, which explains the large frequency of ``GT in Pseudo GT'' and ``€œlow overlap''. For cow, the lower frequency in ``correct localization''€ compared to the other categories is due to the low accuracy of the video tracking algorithm \cite{fanyitube}, which results in sub-optimal object boundaries being transferred to the images; it often tracks multiple cows together instead of a single cow. Finally, it is worth mentioning that for all categories, almost all of our pseudo ground-truth boxes have at least some overlap with a ground-truth box (i.e., the ``no overlap'' frequency is very low). Overall, these results indicate that our approach is able to correctly mine the discriminative regions that correspond to the positive object, and that the transferred video object boundaries using those regions produce good coverage of the positive object to produce accurate pseudo ground-truth boxes.

\subsection{Video track selection accuracy}
\label{subsec:videoaccuracy}
\vspace{-0.04in}

Finally, we evaluate our selection criterion in choosing the relevant object box among the 9 tracks produced by the unsupervised video tracking algorithm~\cite{fanyitube}.  For this, we compute the IOU between the tracked object boxes and the ground-truth boxes on the YouTube-Objects dataset~\cite{prest-cvpr2012}.  Our automatically selected tracks produce a mean IOU of 45.1 over all 10 categories (see Table~\ref{table:video}).  While this is lower than the upper-bound mean IOU of 61.9 (i.e., the max IOU among the 9 proposals in each frame) they are sufficiently accurate to produce high-quality pseudo GT boxes.  Furthermore, since we selectively retrieve a video object box only if it is overlapping with one of the top $n=20$ matching video regions of a discriminative positive region, and then further aggregate those transferred boxes through hough voting, we can effectively filter out most of the noisy transferred tracks (as was shown in Fig.~\ref{fig:pseudo_gt_results}). Overall, we find that~\cite{fanyitube} produces sufficiently good boxes, and our selection criterion is in many cases able to choose the relevant one.  These lead to accurate pseudo GT boxes on the weakly-labeled images.

\section{Conclusions}
\label{sec:conclusion}

We introduced a novel weakly-supervised object detection framework that tracks and transfers object boxes from weakly-labeled videos to images to simulate strong human supervision.  We demonstrated state-of-the-art-results on PASCAL 2007 and 2010 datasets for the 10 categories of the YouTube-Objects dataset~\cite{prest-cvpr2012}.

Our framework assumes that we have a way to track the object-of-interest in videos, so that we can delineate its box and transfer it to images.  This is easier if the object is able to move on its own, but could also work for static objects, as long as the camera is moving.  We plan to investigate this in the future.  Finally, we intentionally trained our detectors using only the weakly-labeled images, in order to make our results comparable to previous weakly-supervised methods.  It would be interesting to explore combining the video tracks with our pseudo GT image boxes for training the object detectors.

\paragraph{Acknowledgements.} This work was supported in part by an Amazon Web Services Education Research Grant and GPUs donated by NVIDIA.

{\small

\bibliographystyle{ieee}
\bibliography{strings,bibs}
}

\includepdf[pages={1-},scale=1]{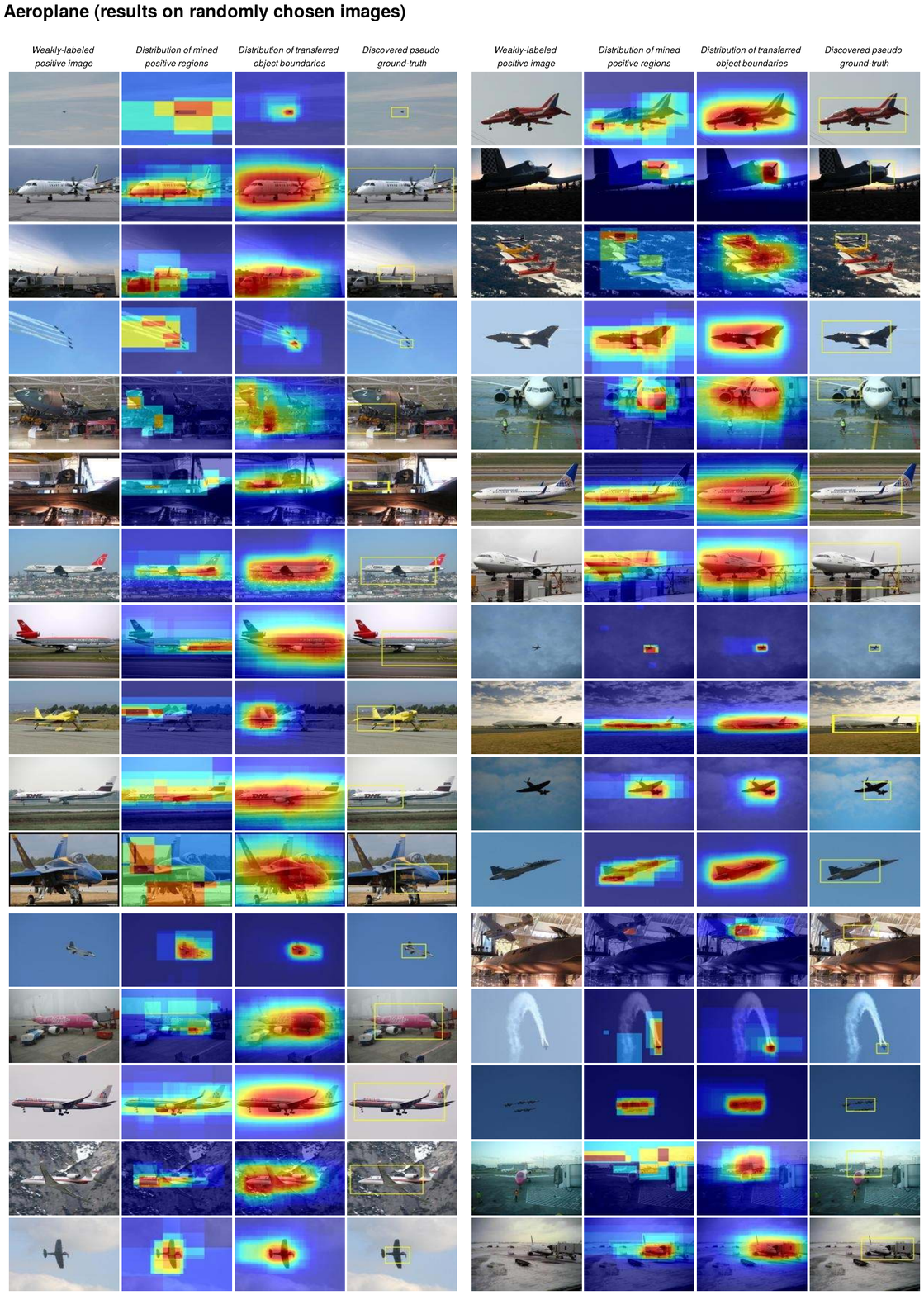}

\end{document}